\documentclass{article}

\usepackage{PRIMEarxiv}

\usepackage[utf8]{inputenc} % allow utf-8 input
\usepackage[T1]{fontenc}    % use 8-bit T1 fonts
\usepackage{hyperref}       % hyperlinks
\usepackage{url}            % simple URL typesetting
\usepackage{booktabs}       % professional-quality tables
\usepackage{amsfonts}       % blackboard math symbols
\usepackage{nicefrac}       % compact symbols for 1/2, etc.
\usepackage{microtype}      % microtypography
\usepackage{lipsum}
\usepackage{fancyhdr}       % header
\usepackage{graphicx}       % graphics
\graphicspath{{media/}}     % organize your images and other figures under media/ folder

\title{A Segmentation Method for fluorescence images without a machine learning approach
\thanks{
\textbf{This research has received funding from the European Union’s Horizon 2020 Framework Programme for Research and Innovation under the Specific Grant Agreement numbers 945539 (Human Brain Project SGA3), Fenix computing and storage resources under the Specific Grant Agreement No. 800858 (Human Brain Project ICEI), and a grant from the Swiss National Supercomputing Centre (CSCS) under project ID ich002. Editorial support was provided by Annemieke Michels of the Human Brain Project. This work was partially financed by Project PO FESR Sicilia 2014/2020 - Azione 1.1.5. - "Sostegno all’avanzamento tecnologico delle imprese attraverso il finanziamento di linee pilota e azioni di validazione precoce dei prodotti e di dimostrazioni su larga scala" - 3DLab-Sicilia CUP G69J18001100007 - Number 08CT4669990220.}} 
}

\author{
  G. Giacopelli \thanks{
  	\textbf{Corresponding author.}}  \\
  Institute of Biophysics \\
  National Research Council \\
  Palermo \\
   \texttt{giuseppe.giacopelli@ibf.cnr.it} \\
 \\
 \And
 M. Migliore \\
 Institute of Biophysics \\
 National Research Council \\
 Palermo \\
 \texttt{michele.migliore@cnr.it} \\
 \And
  D. Tegolo \\
  Department of Mathematics and Informatics \\
  University of Palermo \\
  Palermo \\
  \vspace*{-0.1cm}\\
  Institute of Biophysics \\
  National Research Council \\
  Palermo \\
  \texttt{domenico.tegolo@unipa.it} \\
}

\begin{document}
\maketitle

\begin{abstract}
\emph{Background}:  Image analysis applications in digital pathology include various methods for segmenting regions of interest. Their identification is one of the most complex steps, and therefore of great interest for the study of robust methods that do not necessarily rely on a machine learning (ML) approach.  \emph{Method}: A fully automatic and optimized segmentation process for different datasets is a prerequisite for classifying and diagnosing Indirect ImmunoFluorescence (IIF) raw data. This study describes a deterministic computational neuroscience approach for identifying cells and nuclei. It is far from the conventional neural network approach, but it is equivalent to their quantitative and qualitative performance, and it is also solid to adversative noise. The method is robust, based on formally correct functions, and does not suffer from tuning on specific data sets.  \emph{Results}: This work demonstrates the robustness of the method against the variability of parameters, such as image size, mode, and signal-to-noise ratio. We validated the method on two datasets (Neuroblastoma and NucleusSegData) using images annotated by independent medical doctors.  \emph{Conclusions}:  The definition of deterministic and formally correct methods, from a functional to a structural point of view, guarantees the achievement of optimized and functionally correct results. The excellent performance of our deterministic method (NeuronalAlg) to segment cells and nuclei from fluorescence images was measured with quantitative indicators and compared with those achieved by three published ML approaches.  
\end{abstract}

% keywords can be removed

\section{Introduction} \label{sec:I}
\begin{figure}[h]
	\centering
	\includegraphics[width=0.4\textwidth]{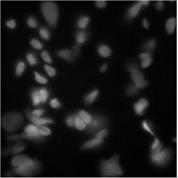}
	\caption{Example of a Complex Image in which some nuclei cannot be clearly distinguished from the cytoplasm.}
	\label{fig:fig1}
\end{figure}

One of the objectives of automatic image analysis is the formalization of methodologies to identify quantitative indicators that characterize elements on pathological slides. The analysis of immunofluorescence images is becoming a fundamental tool for identifying predictive and prognostic elements that can be used to diagnose various pathologies. Cell studies, particularly cancer cell studies, could help to identify predictive parameters to improve patient diagnosis and develop prognostic tests.
The segmentation phase for single cells and nuclei is relatively simple if performed by an expert on cytological images, because, in most cases, cells/nuclei are intrinsically separated from each other. Automated recognition combined with unsupervised and automatic quantitative analysis helps doctors in decision-making and provides cognitive support in the diagnosis of pathologies carried out on slides \cite{Gurcan2009}. This reduces the level of subjectivity, which may affect the decision-making process.
The segmentation process is a necessary first step in obtaining quantitative results on cellular or nuclear images. An accurate and detailed segmentation, in which the single instances of cells/nuclei are highlighted, would provide a valuable starting point for identifying their quantitative characteristics. Incorrect biomedical conclusions may result from the inability of algorithms to separate different and more complex aggregations of cells/nuclei \cite{Hill2007}. For example, complex images, such as those shown in figure \ref{fig:fig1}, can result in incorrect conclusions on specific instances or, in some cases, in disregarding fundamental aggregates to obtain a more accurate diagnosis. Precise segmentation allows highlighting instances on which to focus attention, thus improving the diagnostic process. This process is not the only reason for the development of new methodologies to extract characteristics from digital pathological images, and a better understanding of the pathological processes caused by cellular anomalies will be helpful in clinical and research settings.

In biomedical data analysis, technologies based on big data and machine learning contribute to the identification or prediction of a disease and can improve diagnostic processes and guide doctors towards personalized decision-making \cite{Ristevski2018}. The use of machine learning technologies on biomedical images can hopefully improve clinical processes (diagnostic and prognostic) by reducing human errors. However, in recent years, the use of these methods has highlighted the presence of biases attributable to distortions in the learning dataset or related algorithms that lead to inaccurate decision-making processes. For example, any ML algorithm will only be valid for the data types on which they have been trained; therefore, if there are distortions, these will be reiterated and probably exacerbated by an ML application \cite{Verghese2018}.
Different studies have offered multiple solutions from different perspectives. Some studies have focused on the automated approach analysis of medical images with consolidated tools. This has proven to be rather useful in clinical cases, such as those based on wavelet transforms, which can extract a set of features and discriminate objects \cite{Palermo2019}. Others have developed innovative methodologies that aim to understand the functional and deterministic correctness of the adopted functions. This is important because the availability of formally correct methods, from both functional and structural points of view, guarantees that a method can achieve optimized and functionally correct results.
This study introduces a deterministic method for cell or nuclei segmentation that does not use ML techniques or artificial neural network models and is immune to adversative noise. Our results demonstrate a high degree of robustness, reliability, and computational performance, particularly for noisy images. We believe that this method will significantly improve the ability to understand algorithmic behavior in identifying regions of interest on immunofluorescence images.

\section{Related Works} \label{sec:II}
The design and development of intuitive and efficient methodologies to highlight biological patterns are the mission of a subset of the scientific community. A friendlier framework for managing intelligence modules and complete datasets has grown in the last decade.
Wählby et al. \cite{Whlby2002} presented an article on image analysis algorithms for the segmentation of cells imaged using fluorescence microscopy. The sketch of the method includes an image preprocessing step, a module to detect objects and their merging, and a threshold for the statistical analysis of some shapes describing the features of previous results, which allows the splitting of the objects. The authors declare that the method is fully automatic after the training phase on a representative set of training images. It showed a correct segmentation between 89\% and 97\%.
An interesting study by Rizk et al. \cite{Rizk2014} presented a versatile protocol for the segmentation and quantification of subcellular shapes. This protocol detects, delineates, and quantifies subcellular structures in fluorescence microscopy images. Moreover, the same protocol allows application to a wide spectrum of images by changing some parameter values.
Di Palermo et al. highlighted the use of wavelet transforms applied to the main phases of image analysis in immunofluorescence. They intend to care for the versatility of wavelet transform (WT) and its use in various levels of analysis to classify IIF images to develop a framework capable of performing image enhancement, ROI segmentation, and object classification \cite{Palermo2019}. They reported the success rates (in terms of specificity, sensitivity, and accuracy) of the method for different types of cells to classify them as mitotic and non-mitotic:  CE (98.15, 92.6, 97.9); CS (93.83, 71.1, 91.6); and NU (91.53, 88.1, 91.3).
Such methods are only a few examples of a multitude of image segmentation algorithms introduced in the literature: k-means clustering \cite{Lupascu2010, Basar2020}, graph cuts \cite{Kolmogorov2002, Boykov2001, Liu2019}, active contours \cite{Cootes1995, Soomro2019}, and watershed methods \cite{Roerdink2000, Bieniek2000}. 
An exhaustive dissertation on emerging image segmentation was introduced in \cite{Minaee2020} by Minaee et al., in which approaches based on deep convolutional neural networks and supervised machine-learning methods were introduced to solve the segmentation task as a subfield of the more general classification strategy.
Pan et al. presented an efficient framework in which a simplification of U-Net and W-Net stimulated an original method for nuclei segmentation; they named it attention-enhanced simplified W-net (ASW-Net). Such a method is based on a light network and a cascade structure of network connections; moreover, it hosts a significant connection gate. This infrastructure enables the efficient extraction of features. Furthermore, the adopted post-processing refines the accuracy of the segmentation results \cite{Pan2022}. On the low-resolution images (909 images), the assessment reports that 89\% of the cells were correctly segmented, whereas on the high-resolution images (251 images), 93\% of the cells were correctly segmented; therefore, the method achieved 89–97\% correct segmentation accuracy.
Van Valen et al. investigated multiple cell types; they used a deep convolutional neural network as a supervised machine learning method to present a robust segmentation of the cytoplasm of mammalian cells, as well as of the cytoplasm of individual bacteria. Their assessment demonstrated through a standard index (Jaccard index) that their methodology improves the accuracy of other methods \cite{Valen2016}. Therefore, they assert that deep convolutional neural networks are an accurate method and are generalizable to a diversity of cell types, from bacteria to mammalian cells; then they report a remarkable degree of success in different areas: Bacteria ~0.95(J.I.); Mammalian nuclei ~0.89(J.I.); various Mammalian cytoplasm from 0.77 to 0.84(J.I.).
A remarkable approach is to present a deep learning method to process the simultaneous segmentation and classification of nuclei in histology images. The network HoVer-Net is based on the prediction of horizontal and vertical maps of nuclear distances from their centers of mass to detach the clustered nuclei. Graham et al. argued that the HoVer-Net network achieves at least the same performance as introduced by several recently published methods on multiple H\&E histology datasets \cite{Graham2018}. Moreover, they declare to experiment with their methodology on different exhaustively annotated datasets and perform instance segmentation without classification with a DICE percentage ranging from 0.826 to 0.869.
The Residual Inception Channel Attention-Unet, an Unet-based neural network for nuclei segmentation, was proposed in \cite{Zeng2019} (RIC-Unet). The authors include techniques of residual blocks, multi-scale, and channel attention mechanisms in RIC-Unet to achieve a more precise segmentation of nuclei. The effectiveness of this approach was compared to traditional segmentation methods and Neural network techniques and tested on different datasets. They reported three quantitative indices (Dice, F1-score, Jaccard) ranging from 0.8008 to 0.7844, 0.8278 to 0.8155, and 0.5635 to 0.5462, respectively.
Machine learning has a metamorphic impact on image segmentation; thus, the scientific community considers it a powerful tool for analyzing biomedical images. However, the training phase, tuning parameters, and statistical learning conditions are ML, AI, and DL processes. In contrast, our method does not require any fine-tuned parameters nor has any training phase, and it is robust in terms of adversative noise. Then, it is disconnected from standard Artificial Intelligence approaches and linked to more actual neuronal intelligence.

\section{Methods and Data} \label{sec:III}
This section presents some methods for comparing the proposed methodology with a set of datasets to validate its effectiveness. Section \ref{sec:III} describes some machine learning techniques used to compare our method. Datasets with  cells and nuclei are also described. In addition, a brief description of the Otsu algorithm, wavelet transform, and active contour models used in the preprocessing phase is provided in Section \ref{sec:III-II}. 

\subsection{Machine Learning Approaches} \label{sec:III-I}
In recent decades, neuronal networks have proven to be a reasonable solution to various computational problems \cite{Deng2014}, \cite{Khan2019}, \cite{Tajbakhsh2016} and have achieved considerable success in the segmentation and classification of objects present in digital images \cite{Alzubaidi2021}, \cite{Lavitt2021}. The basic technique consists of automatic learning from a training set with a multilevel hierarchical strategy; in some cases, this functionality is invariant with respect to small or large variations in the learning samples, producing relevant results \cite{Khan2019}, \cite{Liang2017}.
A Convolutional Neural Network (CNN) can be identified among the most performing neuronal networks, consisting of pairs of convolutional levels coupled to connected levels. The fundamental components of a CNN can be summarized as follows: convolutive filters operating within convolutional levels in which the goal is to produce a map of features from input images; grouping functions in which the outputs of the convolutional levels converge and in which maximum values are selected; functions for the assignment of the probabilities in which the data coming from the grouping functions are allowed, and in which the probabilities that the input data belong to a specific class are assigned \cite{Pan2018}. The following section provides a brief description of some of the best and most popular N.N. for image segmentation. In addition, this section provides a comparison with our method.

\subsubsection{U-Net}
U-Net is a deep learning network \cite{Ronneberger2015} for image processing. The idea is to scale down the information of the input image through convolution layers and then scale up the information through transposed convolutional layers to obtain an image with the exact resolution of the original image with the information of the semantic segmentation in each pixel. U-Net is the most popular segmentation algorithm for several reasons:

\begin{itemize}
	\item The architecture is so simple that it can be applied to many medical imaging segmentation tasks \cite{Siddique2020}; 
	\item  Even if the network is deep, it can be trained in a short time requiring low computational resources (the U-Net used in this study \cite{Ronneberger2015} was trained on a GPU RTX 2070 with 8 GB of VRAM in approximately 1 hour); 
	\item This network requires few computational resources for prediction, and it is very fast in forwarding time \cite{Wu2019}.
\end{itemize}

The main disadvantage of this architecture is that the segmentation of the cells/nuclei task returns a binary label and cannot separate single cells/nuclei by default. In this work, we use the Keras/Tensorflow implementation available at  \url{https://github.com/zhixuhao/unet} inspired by \cite{Ronneberger2015} with grayscale images at a resolution of $512 \times 512$. 

\subsubsection{KG Network}
The Keypoint Graph Network (since now KG Network) is a neural network based on the concept of Keypoint Graph \cite{Yi2019}. The network first applied ResNet34-based feature extraction. Then, the network layers identify some points (called keypoints) that discretize the input image, and the collected keypoints are processed to extract the bounding boxes of the cells/nuclei. Finally, the bounding boxes of the cells/nuclei are taken as the input for the final layers that extract the cells/nuclei masks. This network (in contrast with the previous U-Net) provides object-by-object segmentation. However, it has a more considerable forward time than U-Net and a training time of approximately 4 hours on the same machine with a GPU RTX 2070 with 8 GB of VRAM. In this study, we used a PyTorch implementation publicly available at the website  \url{https://github.com/yijingru/KG_Instance_Segmentation} based on Ref. \cite{Yi2019}, with grayscale images at a resolution of $256 \times 256$.

\subsubsection{R-CNN}
The mask R-CNN uses a Region-based Convolutional Neural Network (R-CNN) \cite{He2017} to extract the masks of single cells and nuclei. This network has a more significant forward time than the previous networks, and it requires a massive quantity of VRAM to be trained. In fact, we trained it on a cloud node with a GPU NVIDIA K80 with 24 GB of VRAM for 2 hours and 30 min. For this study, we used a Tensorflow/Keras implementation publicly available at \url{https://github.com/matterport/Mask_RCNN} with grayscale images at a resolution of $256 \times 256$.

Fluorescence (F.) is a technique used to detect specific biomolecules within a tissue or cells/nuclei using specific antibodies that contain fluorescent dyes \cite{Im2018}. There are different F-techniques. A distinction can be made between direct and indirect F \cite{Im2018}: in direct F, the antibody carrying the fluorophore (the fluorescent substance) binds directly to the biomolecule, whereas in indirect F, the antibody carrying the fluorophore binds to other antibodies or molecules that are directed against the biomolecule of choice, thus binding indirectly to it. Our discussion focuses more on the results of F. than on the methods involving this procedure. The result of F. is a microscopy image that shows the fluorescence of the area bound by the antibody, in contrast to the darker regions not bound by the antibody. The details of the microscopes used for each dataset are specified in the following sections. The proposed algorithms work with grayscale images; subsequently, we will assume a preprocessing step that extracts such information from each dataset. This assumption generally holds \cite{Kromp2020}, but it is not true for any dataset. For example, some datasets are RGB, and they contain most of the information contained in the green channel \cite{Nigam2015} or even on the blue channel \cite{Gunesli2020}. Therefore, in the proposed analysis, it is implied that the starting point is a grayscale version of every dataset with as much information as possible. 

\subsection{Deterministic Approaches} \label{sec:III-II}
Several approaches have not used machine learning for object segmentation. However, most of them are characterized by a single feature: they work for a combination of parameters depending on the dataset (in the worst cases, only on a single image), but this combination must be manually found by trial and error. For this reason, these approaches are described below to have a wide perspective on the current state of the art. Although they are not fully tested in this study, a few of these approaches will be used in our pipeline.

\subsubsection{Otsu's Method}
The first segmentation approach was Otsu’s method \cite{Otsu1979}. It is based on the simple idea that if a white object is placed on a dark background, then there must be two peaks in the grayscale pixel value histogram \cite{Stockman2001}: the first one is the most common background color, and the second one is the most common object color. Thus, the idea behind Otsu’s method is to find a grayscale value $T_{Otsu}$ such as defining two classes of pixels as

\begin{itemize}
	\item The class (class 0) of pixels with grayscale value smaller than $T_{Otsu}$;
	\item The class (class 1) of pixels with grayscale value greater than $T_{Otsu}$.
\end{itemize}

Otsu’s method iterates the threshold $T_{Otsu}$ from 0 to 255 to determine the value that maximizes the variance. It is a powerful tool because it does not require any tuning by the user. However, it has many disadvantages that render it impractical for real-world images. The most common underlying assumption is that a perfectly bright object is placed on a completely dark background. This rarely occurs in real-world images \cite{Kittler1985}. If the previous hypothesis holds, the color histogram is a perfect bimodal distribution; however, real-world images are noisy, making Otsu’s segmentation difficult. Another common problem is that when the colors of the object and background are very distinguishable, the object is very small in comparison to the background \cite{Lee1990}. In this case, the peak of the object pixels in the histogram exists, but is too small to be relevant in terms of variance. This case will be recurrent in the proposed model and solved using our neuronal agents. In conclusion, Otsu’s method is very powerful, but when used alone, it can lead to poor performance.

\subsection{Watershed}
One of the greatest limitations of Otsu’s method is that it returns a binary, where every pixel indicates class 1 or class 0. However, in many applications, it is not sufficient to identify the pixel class, and it is also required to discriminate between different objects of the same class inside the image. A classic example of this is the analysis of cellular microscopy images. In this case, the user is not only interested in where cells/nuclei are placed, but they could also be interested in distinguishing different cells/nuclei to count or classify them. Therefore, a watershed \cite{Beucher2009} was introduced. This transform can be used to distinguish homogeneous objects based on the gradient of the image. Several versions of this algorithm have been proposed \cite{Kornilov2018}. We consider the most common, that is, Meyer’s watershed \cite{Meyer1992}, implemented in the open-source framework OpenCV \cite{OpenCV}. The transform starts with a set of markers established by the user (most of the time, it is extracted in an automated manner using mathematical morphology). The algorithm performs a “flooding” of the image in order to find the optimal “basins” using the following procedure:

\begin{itemize}
	\item The markers are initialized with the user’s input;
	\item The neighboring pixels of a marked pixel are inserted into the queue with a priority proportional to the gradient modulus of the image of the inserted pixel;
	\item The pixel with the highest priority is extracted. If the surrounding pixels marked have the same label, the pixel is marked with this label. All the surrounding pixels that are not yet marked are inserted in the queue;
	\item Return to step 2 until the queue is empty.
	
\end{itemize}

Watersheds are one of the simplest algorithms for splitting purposes, and with good quality images, they perform very well. However, this method has several disadvantages. Often, the initial markers are selected starting from Otsu’s thresholding, which means that these implementations could be affected by the same problems as described in the previous section. Another disadvantage is that in noisy images, a watershed can be affected by oversplitting \cite{Kornilov2018}, meaning that there are more clusters than expected. In addition, for very close and merged objects (such as cell/nucleus clusters), mathematical morphology methods can fail to separate single objects. For these reasons, the proposed model has some preprocessing steps in the markers’ individuation and post-processing steps of the watershed masks to obtain optimal results.

\subsubsection{Active Contour Model}
The snake or active counter model was widely used in this study. However, our Neuronal Agents lead to a better generalization without parameter tuning; thus, they replace the snake classic functional. The following section describes some aspects of a general snake to better understand the entire process.  
The idea of the active contour module is to build a discrete contour made by key points (often called a snake) aimed at minimizing a line function. To define this line functional, three energies associated with ACM must be defined \cite{Kass2004}. The first is $E_{int}$ which is the internal energy functional defined to keep the ACM as rigid as needed and to avoid excessive ACM shrinking. 
For the second energy, that is the image energy functional defined as, 

$$E_{img}= w_{line} E_{line}+w_{edge} E_{edge}+w_{term} E_{term}$$

where  $E_{line}$ from the grayscale values of the image , $E_{edge}$ from its gradient, the energy $E_{term}$ takes into account the normal derivative of the image gradient, and the constants $w_{line}, w_{edge}, w_{term}$ must first be defined. Therefore, A.C.M. minimizes the functional calculated along the contour.
$E_{con}$ is an additional component used to guide the user to the process by adding constraints to the run \cite{Kass2004}. 
Its main critiques towards this kind of algorithm are:
Problems involving the internal functional for complex shapes \cite{Ciecholewski2016};
Convergence of the algorithm to local minima \cite{Akbari2021}.
The first class of problems is caused by the fact that A.C.M. uses $E_{img}$ to attract the snake towards the edge and $E_{int}$ to avoid the collapse of the snake at a single point, keeping it as smooth as possible. The second type of problem is caused by the process of minimization that can converge to a local minimum (caused, for example, by a source of noise), leading to poor convergence to the actual contour. These kinds of problems imposed the definition of a new contour model implemented by our Neuronal Agents, which is immune to parameter tuning.

\subsection{Datasets} \label{sec:III-III}
Here it is a brief description of the dataset involved in the experiments.

\subsubsection{Neuroblastoma Dataset}
The first dataset was the dataset introduced in \cite{Kromp2020}. It was composed of four samples of tumor tissue and four samples of the bone marrow of Neuroblastoma patients. The dataset was created with the aid of the Children’s Cancer Research Institute (C.C.R.I.) biobank (EK.1853/2016) to establish a benchmark for experiments on automatic cell nuclei segmentation. The dataset consisted of 41 training images and 38 test images in jpg format with a resolution variable of approximately $1200 \times 1000$ F. cells/nuclei. The images are already in grayscale, which means that brighter zones are white and darker zones are black, so they do not require preprocessing. Segmentation was performed manually by the authors \cite{Kromp2020}, and it distinguished different cells/nuclei. Segmentation was stored in text-based files. The dataset has proven to be a hard enough benchmark to test models on real-world images \cite{Kromp2021}. For these reasons, it is considered the main benchmark of this study.

\subsubsection{NucleusSeg Dataset}
The second dataset was introduced in \cite{Koyuncu2018} and then used in \cite{Gunesli2020}. It is composed of 61 RGB images with a resolution variable of approximately $1000 \times 700$ cancer cells/nuclei taken from the Huh7 and HepG2 regions. Koyuncu et al. acquired images under a Zeiss Axioscope fluorescent microscope with a Carl Zeiss AxioCam MRm monochrome camera with a 20× Carl Zeiss objective lens. For the Hoechst 33258 fluorescent dye, a bisbenzimide DNA intercalator can be observed in the blue region upon UV region excitation. Hoechst 33258 dye was excited at 365 nm and emitted blue light (420 nm) was acquired \cite{Koyuncu2018}. Therefore, the bright color in this dataset was not white but blue. Therefore, this dataset was preprocessed to extract the blue channel of the image to obtain a grayscale image representative of fluorescence. This dataset was not used to train the model because it did not include a variety of Neuroblastoma datasets. However, it was used as a test set to examine the generalization capacity of the algorithms. 

\subsubsection{ISBI 2009 Dataset}
The third dataset has been introduced in \cite{Coelho2009} and then used and is composed of 46 grayscale images with a resolution variable of about $1350 \times 1000$ of U2OS cells, created explicitly for computer vision benchmarks. To the best of our knowledge, the authors don’t declare the microscope used and the IF methods involved. Moreover, the dataset has been segmented by hand by authors, making the handmade segmentation in GIMP files publicly available. Then, the dataset was preprocessed to transform data images and the author’s segmentation into explicit images.

\section{Proposed Method} \label{sec:IV}

\subsection{Why a new approach is needed}
The previous algorithms were all based on Artificial Neural Networks, and they performed well under optimal conditions. However, literature shows that Deep Neural Networks (DNN) can be affected by adversarial attacks \cite{Goodfellow2014}. An adversarial attack is a small perturbation (often called adversarial noise) introduced and tuned by a machine-learning algorithm to induce misclassification of the network. Segmentation neural networks are not immune to these types of attack. For example, in \cite{Bar2021}, it was proved that an adversative attack can fool the ICNet \cite{Zhao2017}, which provides semantic segmentation by controlling an autonomously driving car. The changes in the adversative algorithm applied to the input image are so subtle that a real-world light imperfection or a camera sensor not working properly can recreate the adversative pattern, leading to an accident. This unlucky (but possible) scenario reveals the second major problem of these algorithms: they are black boxes, and in many cases, after an accident, the best solution is to train the network again, hoping that no similar issues will affect the network in the future. Therefore, the concept of Explainable Artificial Intelligence (XAI) is becoming increasingly popular. In brief, XAI is an algorithm that can be designed to determine its actions and eventually correct them. This is of crucial importance for the real-world application of deep learning in fields such as robotics, automation, and medicine because an XAI can be fixed after an error, and a human can eventually be accountable for its errors. These assertions induce the reformulation of new methods in which DNN performance is guaranteed, and the correct approach of XAI must be considered. Therefore, our strategy defines a new method to host formal correctness and DNN performance, as well as in the presence of adversative noise.

\begin{figure}[h]
	\centering
	\includegraphics[width=0.4\textwidth]{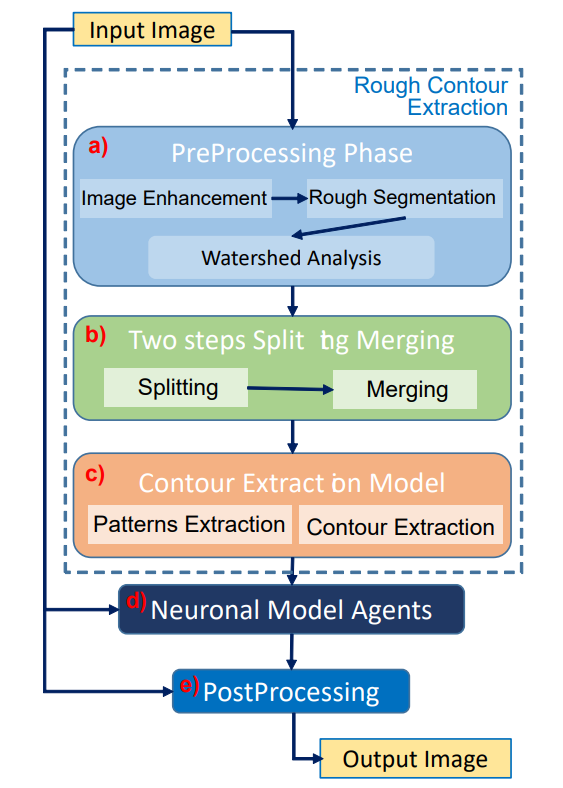}
	\caption{Sketch of our deterministic method (NeuronalAlg), sections a) preprocessing module extracts a rough segmentation of input image; b) split and merge module as a first step to improve the previous segmentation; c) last step of the rough segmentation; d) main core of the whole task: Neuronal Method. This module improves the segmentation with a Neuronal agent; e) post processing phase to extract the binary mask.}
	\label{fig:fig3}
\end{figure}

\subsection{The proposed model}
We propose an X.A.I. algorithm based on mask transformation using neuronal-based agents. The algorithm comprises six main parts, as shown in figure \ref{fig:fig3}:

\begin{enumerate}
	\item Preprocessing phase: prepares the images for the next steps.
	\item Watershed analysis: splits cells/nuclei using the well-known watershed transformation \cite{Kromp2020}.
	\item Two steps of splitting and merging: improves watershed separation;
	\item Extraction Phase: From the cells/nuclei masks, the contour was extracted to run the neuronal model.
	\item Neuronal Method: manipulate the contour of the mask using neuronal agents;
	\item Post-processing phase: Some thresholding algorithms improve mask precision.
\end{enumerate}

The main goal of the proposed algorithm is to segment bright (colored) cells/nuclei on a dark background in the grayscale images. However, the dataset proposed in the previous sections consists of blue or green RGB images. The underlying hypothesis is that the dataset was processed to extract the brightness of cells/nuclei to obtain a grayscale image suitable for further processing. For example, Hoechst staining data have been extracted from the dataset NucleusSegData \cite{Gunesli2020}, which was obtained using U.V. light, and the color of such images was blue. Therefore, in this case, the blue channel was extracted from the image to obtain a grayscale image with the desired properties. So in the first phase all the cells are taken to the same color encoding (white cell, black background). Moreover using a series of lowpass gaussian filter is extracted a smoothed version of the image for the next steps.

To this smoothed version of the image is applied a watershed transform to make a rough subdivision of the cells. This procedure make an approximated subdivision of the cells, however it is not enough to obtain results comparable to other state of the art methods. For this reason we continue in our routine using a Split-Merge method. 

This method is based on the assumption that the cells size inside the image is uniform. Defined for every cluster $i$ the area of the cluster $A_i$, can be compute the average cluster area as
$$ A_{avg} = \frac {\sum_{i=0}^ {N_{clusters}} A_{i}} {N_{clusters}} $$

Then if a cluster $j$ has an area significantly higher than the current average cluster area is split in a number of subcluster as close as possible to the number
$$ r_j = \frac{A_j} {A_{avg}} $$
using a recursive algorithm inspired to Otsu's thresholding. This Split-Merge procedure is applied twice. The result of the previous steps is a collection of mask containing the cells, but with countours not perfectly matching the the cells border.

\begin{figure}[h]
	\centering
	\includegraphics[width=0.4\textwidth]{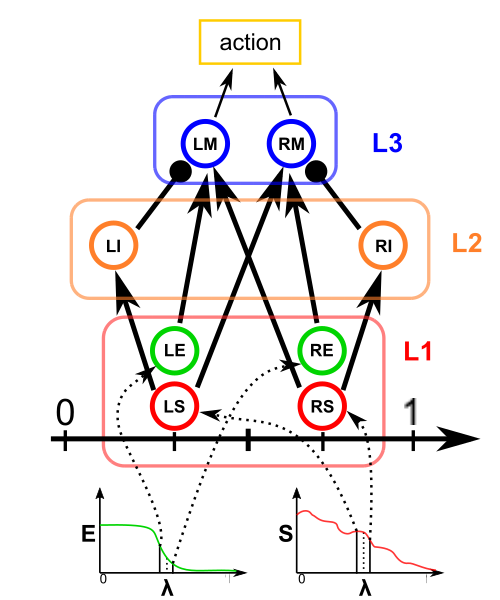}
	\caption{Sketch of Neuronal agent.}
	\label{fig:fig4}
\end{figure}

Now every cell mask previously computed is subdivided into 30 radial bins, and the mean radius is calculated for each bin. A collection of 30 points following the contour of the mask was obtained from the bin angle and the average radius. This phase is shown in Figure \ref{fig:fig3}c as a contour-extraction module.
Then, the generated contour provides the idea of the displacement of an object, but it is not sufficiently precise to compete with a state-of-the-art model. Therefore, each point of the contour is associated with a neuronal agent that moves along the line starting from the center of the object and passing through the original contour point. This agent is then a 1D agent that can move in the direction of the center or away from the center. The motion of the agent is controlled by the simple idea that the agent is repulsed by high grayscale values of the input image but is attracted by the segmentation mask previously computed. These conditions create a system that converges to the position of equilibrium, which is the actual contour of the object. The agent is a neuronal network built into NEST \cite{NEST} with eight neurons distributed in three layers (see figure \ref{fig:fig4}). The first layer (L1) is the input layer and is composed of four neurons: the LS, LE, RE, and RS. To calculate the intensity of the current to which they have subjected these neurons, we need to calculate the images $gray$ as an equalized version of the gray scale input image, $mask$ that is a Gaussian smoothing of the binary mask associated to the cell mask computed and $mask_{filt}$ that is defined as
$$ mask_{filt} = \frac{2\; mask + 0.1}{1+0.1} $$

In conclusion for each pixel $p=(p_x,p_y)$ it is possible to calculate

$$ E\left(p_x, p_y\right) = 3000 * \left( \frac{\sum_{i= p_x -sd}^{px+sd} {\sum_{j= p_y -sd}^{py+sd} {mask_{filt} (i,j)} } } {2 \; sd^2} \right)$$

$$ S\left(p_x, p_y\right) = 3000 * \frac{3}{8} * \left( \frac{\sum_{i= p_x -sd}^{px+sd} {\sum_{j= p_y -sd}^{py+sd} {gray (i,j)} } } { sd^2 \; gray_{avg}} \right)$$

where $sd$ is the closer even number to $10\; sf$ (with $sf = (image_{Height}+image_{Width})/2220$), the 3000 value is the stimulation intensity, and 3/8 is an empirical value fixed for all experiments. Now, if $p_0$ is the center of the object and $p_c$ is the contour position, for each agent, the coordinate $\lambda$ is introduced such that the position of the agent can be expressed as
$$p_a(\lambda) = \lambda p_c + (1-\lambda) p_0$$

We defined the stimulation current for neurons LS ($S_{LS}$), RS ($S_{RS}$), LE ($E_{LE}$), and RE ($E_{RE}$) as
 
$$S_{LS} (\lambda)=S(p_a (\lambda-0.2))$$
$$S_{RS} (\lambda)=S(p_a (\lambda+0.2))$$
$$E_{LE} (\lambda)=E(p_a (\lambda-0.2))$$
$$E_{RE} (\lambda)=E(p_a (\lambda+0.2))$$

The neuron LS has an excitatory synapse on the neuron LI and the neuron RM; in contrast, the neuron RS has an excitatory synapse on the neuron RI and the neuron LM. The LE and RE neurons have excitatory synapses on the LM and RM neurons, respectively. The role of this first layer is to “perceive” the image configuration and send it to the next layers. The second layer (L2) is made up of two inhibitory neurons, LI and RI, which have an inhibitory synapse on the neurons LM and RM. Their role is to inhibit the motor neurons LM and RM when the LS and RS neurons are activated. Neurons LM and RM, included in layer 3, are responsible for the motion of the agent. The speed of the agent is calculated as
$$speed = 0.06 \left(Spikes_{RM}-Spikes_{LM} \right)$$

where $Spikes_{RM}$ and $Spikes_{LM}$ are the numbers of spikes in the simulation window (5 ms). Following the previous steps, each agent reaches the convergence point. 
After this, the agent mask is post-processed using Otsu segmentation \cite{Ciecholewski2016} inside the masks, which has good performance because the content inside the object mask is bimodal. However, some cells/nuclei were still clustered, and for this reason, the last splitting and merge cycle was performed. This last splitting differs from the previous splitting because it is based on the distance transform L2 \cite{Ye1988}.

\subsection{Evaluation criterion and metrics}
Hereafter, only the binary segmentations of the methods discussed will be evaluated, even if some of them return object-by-object segmentation. In general, a bit of binary object segmentation can be called Positive if the segmentation shows that there is an object on it, or Negative if not. To evaluate the goodness of binary segmentation, there are many metrics with slightly different interpretations, but every metric proposed is based on four sets: True Positives (T.P.), True Negatives (T.N.), False Positives (F.P.), and False Negatives (F.N.). The T.P. set is the set of pixels Positive for the ground truth and the prediction. Similarly, the T.N. set is the set of pixels Negative for the ground truth and the prediction. The F.P. set is the set of pixels that are Positive for the prediction and Negative for the ground truth. Analogously, the F.N. set is the set of pixels Negative for the prediction and Positive for the ground truth. The metrics used are:
\begin{itemize}
	\item Intersection over Union (IoU): This is defined as $\frac{TP}{TP+FP+FN}$ and is one of the most balanced metrics.
	\item F1-score: is defined as  $\frac{2 TP} {2 TP+FP+FN}$ and can be proven to be almost proportional to the IoU.
	\item Accuracy is defined as  $\frac{TP+TN}{TP+FP+FN+TN}$ and is one of the most popular metrics of machine learning. However, in object segmentation tasks, this metric can be biased in cases of few sparse cells/nuclei; in these cases, the number of negative pixels can be much greater than the number of positive pixels. This means that even if the prediction is fully negative (every pixel is negative), if the ground truth ratio P/N tends to 0, then the accuracy tends to 1.
	\item Sensitivity: is defined as  $\frac {TP}{TP+FN}$ and can be biased if the ground truth ratio N/P tends to zero.
	\item Specificity: is defined as  $\frac {TN}{TN+FP}$ and can be biased if the ground truth ratio P/N tends to zero.
\end{itemize}
	
\section{Results} \label{sec:V}
We tested NeuronalAlg against two datasets with and without adversative noise and ground truth as baselines to evaluate the performance and goodness of our method. Tables \ref{tab:t1}-\ref{tab:t15} show that the method reports equivalent results of the NN and overperforms them in the presence of adversative noise (Fast Gradient Sign Method). Therefore, NeuronalAlg achieves notable performance in cells/nuclei segmentation, and, contrary to NN, it does not require any training phase to evaluate it on other datasets.

\begin{figure}[htp]
	
	\centering
	\includegraphics[width=0.95\textwidth]{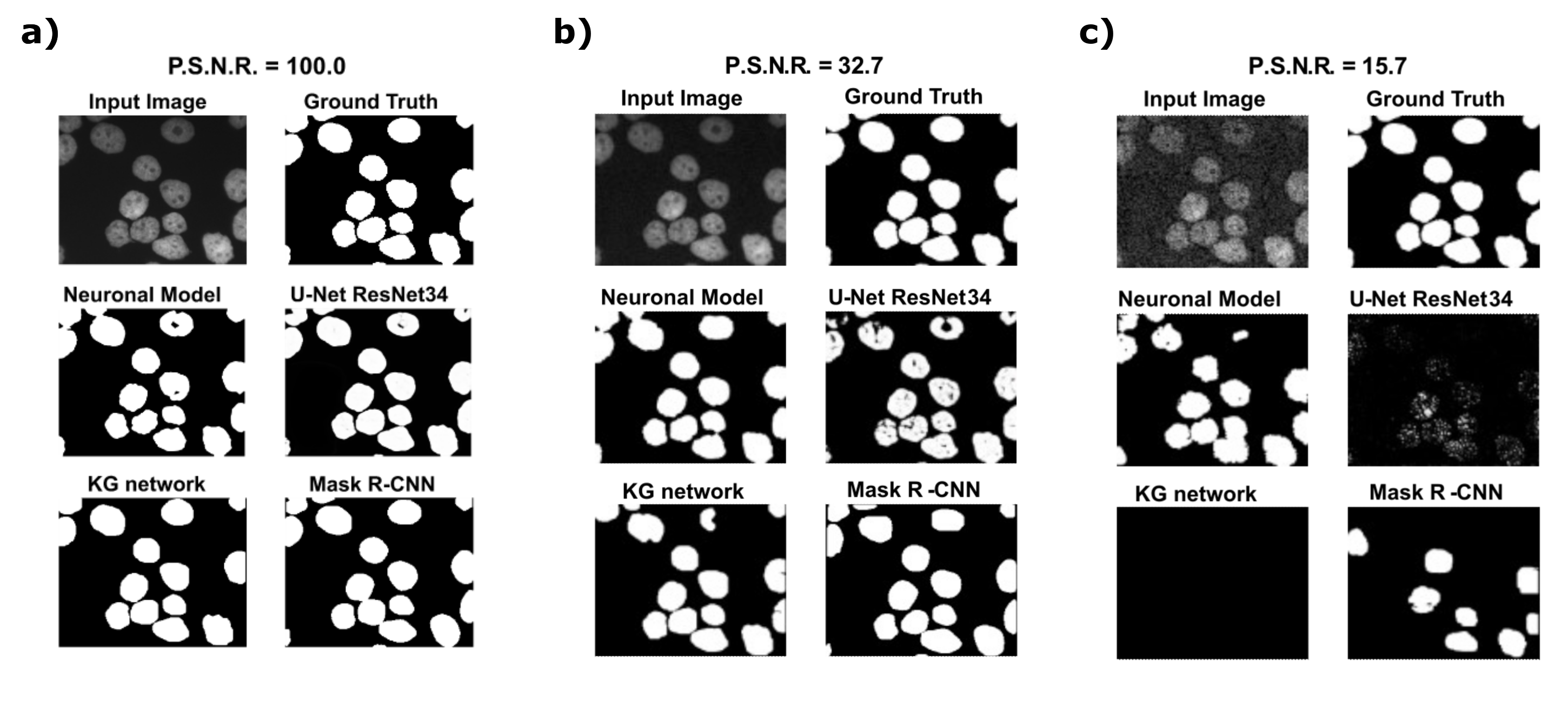}\hfill
	
	\caption{Results for an input image: (a) without noise, (b) with a 32.7 P.S.N.R and (c) with a 15.7 P.S.N.R. }
	\label{fig:fig5}
	
\end{figure}

\subsection{Adversative noise}
Using the Fast Gradient Sign Method (F.G.S.M.) methodology in \cite{Zhao2017}, we performed an adversative attack towards the network U-Net. The F.G.S.M. epsilon value was set to 0, 0.01, 0.025, 0.05, 0.1, and 0.2. figures \ref{fig:fig5}a-c show the sample images related to the Neuroblastoma dataset. U-Net, in the absence of noise, performed very well (figure \ref{fig:fig5}a, second row). However, in the presence of a slight noise (figure \ref{fig:fig5}b, second row), the U-Net segmentation starts to exhibit large holes in the cells/nuclei, and the phenomena become worse if the noise becomes consistent (figure \ref{fig:fig5}c, second row). In these cases, the segmentation almost disappeared and no object was shown. One of the most common critiques against adversative noise is that the M.L. model that generates it is trained on the network and is strongly focused on the analyzed network. Then, the images obtained by the F.G.S.M trained on the U-NET were evaluated by the KG network and mask R-CNN network. The result is that the KG network is very resistant to intermediate noise (figure \ref{fig:fig5}b, third row), but often returns no segmentation with high noise values (figure \ref{fig:fig5}c, third row). The mask R-CNN showed good performance under every noise condition, even if some cells/nuclei were lost with strong noise (figure \ref{fig:fig5}c, third row). The best performance was achieved using the proposed model. Indeed, the segmentation is steady for low and intermediate noise (figure \ref{fig:fig5}a and 5b, third row) and exhibits only a few holes if the noise is strong (figure \ref{fig:fig5}c, third row).

\begin{figure}[h!]
	\centering
	\includegraphics[width=0.75\textwidth]{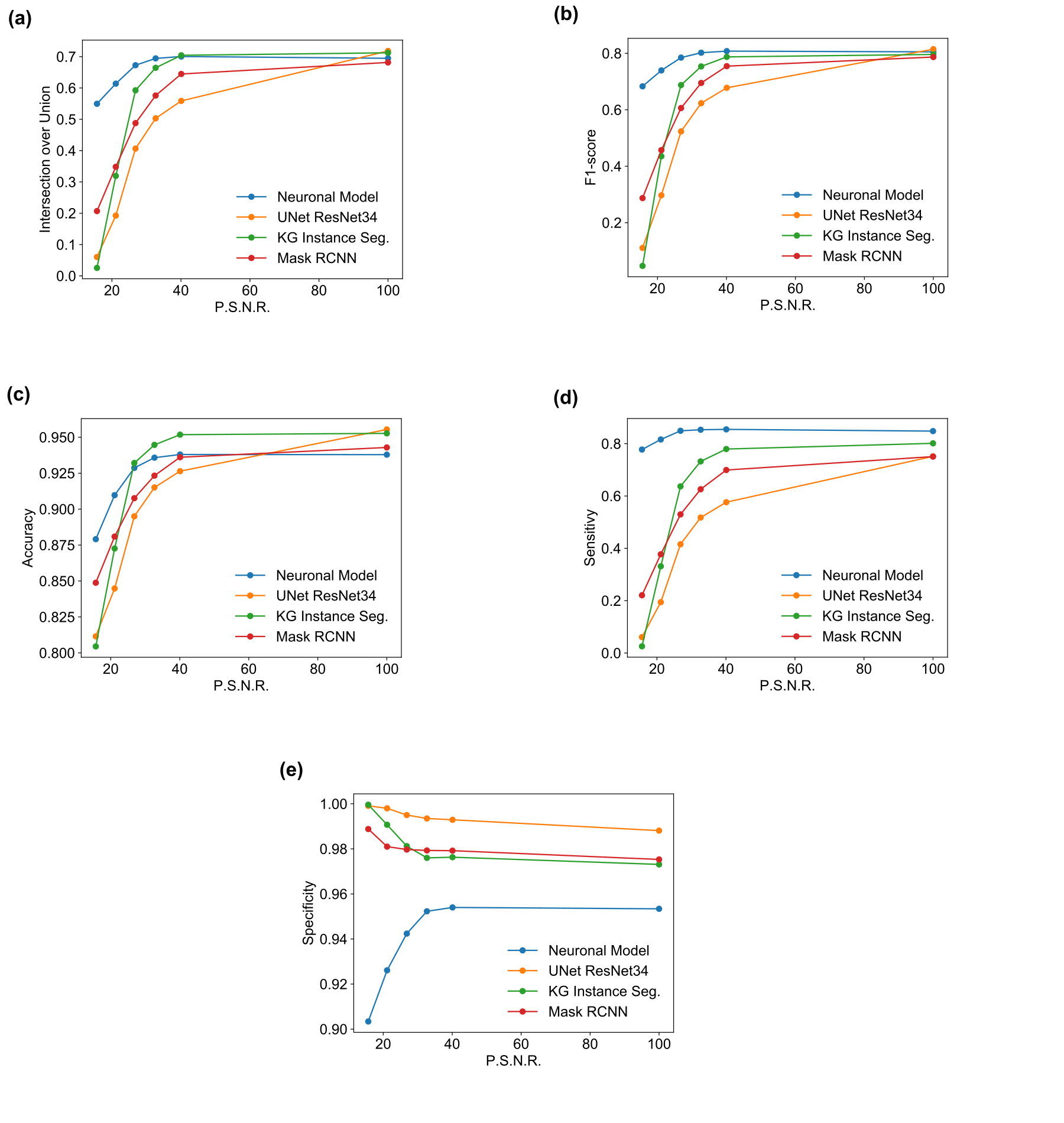}
	\caption{performances in terms of IoU (a), F1-score (b), Accuracy (c), Sensitivity (d) and Specificity (e) for each algorithm on Neuroblastoma dataset.}
	\label{fig:fig6}
\end{figure}

\subsection{Results on Neuroblastoma Dataset}
Judging algorithms based on a few images are not a good practice, so we conducted a quantitative investigation of the results of the algorithms with adversative noise with PSNR values of 100.0, 40.1, 32.7, 26.9, 21.1, and 15.7. The algorithms were evaluated using a test set of 38 images. Each image was compared with the ground truth to compute the metrics of intersection over union, F1-score, Accuracy, Sensitivity and Specificity. The results are plotted in figure \ref{fig:fig6} and summarized in table \ref{tab:t1}-\ref{tab:t5} (in the Appendix). It can be observed from figure \ref{fig:fig6}a and table \ref{tab:t1} that in terms of IoU for P.S.N.R. 100.0 (no noise), the U-Net has the best performance with an IoU of 0.718, followed by the KG network with an IoU of 0.712, the proposed Neuronal algorithm with an IoU of 0.708, and finally the Mask R-CNN with an IoU of 0.682. However, by adding adversative noise, the situation changes because U-Net shows the steepest descent, but these results can be explained by the fact that adversative noise has been created ad hoc for this network. The KG network seems to maintain good performance for a low amount of adversarial noise, but with P.S.N.R. less than 25, the IoU drops to less than 0.32. The other algorithm outperforms the mask R-CNN for low noise, but by adding a strong noise, IoU has a slower decay than the other DNNs, even if the maximum adversarial noise reaches an IoU of 0.21. The proposed model outperformed the DNN in terms of noise resistance because with P.N.S.R. 15.7 (the highest noise evaluated) reaches the minimum IoU value of 0.55, which is more than twice that of all other algorithms.
The F1-score (figure \ref{fig:fig6}b and table \ref{tab:t2}) has a behavior equivalent to that of the IoU, and all the considerations done for the IoU still hold for the F1-score.
Accuracy (figure \ref{fig:fig6}c and table \ref{tab:t3}) and all other metrics must be carefully analyzed. If U-Net, KG network, and mask R-CNN produce false classifications, they are more prone to false negatives (F.N.). In contrast, the proposed neural algorithm is more prone to produce a false positive (F.P.) in the case of false classification.
This event translates into the observation that the DNNs usually have more background (negatives) than the ground truth, and they are under-segmented because cells/nuclei parts are cut by the algorithm being classified as background. On the other hand, the proposed model has shown over-segmenting because part of the background has been classified as cells/nuclei parts. This simple observation slightly conditioned the Accuracy because, in the under-segmenting algorithms, the number of T.N. is more significant than in an over-segmenting algorithm. However, the entire graph confirms that the proposed model is more resistant to adversarial noise.
The previous event became clear in terms of Sensitivity (figure \ref{fig:fig6}d and table \ref{tab:t4}) and Specificity (figure \ref{fig:fig6}e and table \ref{tab:t5}). Indeed, the proposed Neuronal algorithm outperforms the DNN algorithms in terms of Sensitivity (because in this case, F.N. is very small), whereas the DNN algorithms outperform the proposed algorithm in terms of Specificity (because in these cases, F.P. is very small). However, the Specificity case is very curious because it seems that the larger the noise, the larger the Specificity. The answer is hosted in figure \ref{fig:fig5}a-c, in which many DNN predictions with strong noise are almost Positives-free (black segmentation); in this case, T.P.=0 or F.P.=0, which means that it holds Specificity= 1. Regardless of the value of T.N.. 
In these cases, with stronger noise, positive-free segmentation is more likely to be achieved with higher Specificity. For these reasons, more balanced metrics (such as IoU and F1-score) are preferred for the interpretation of results.

\begin{figure}[h!]
	\centering
	\includegraphics[width=0.75\textwidth]{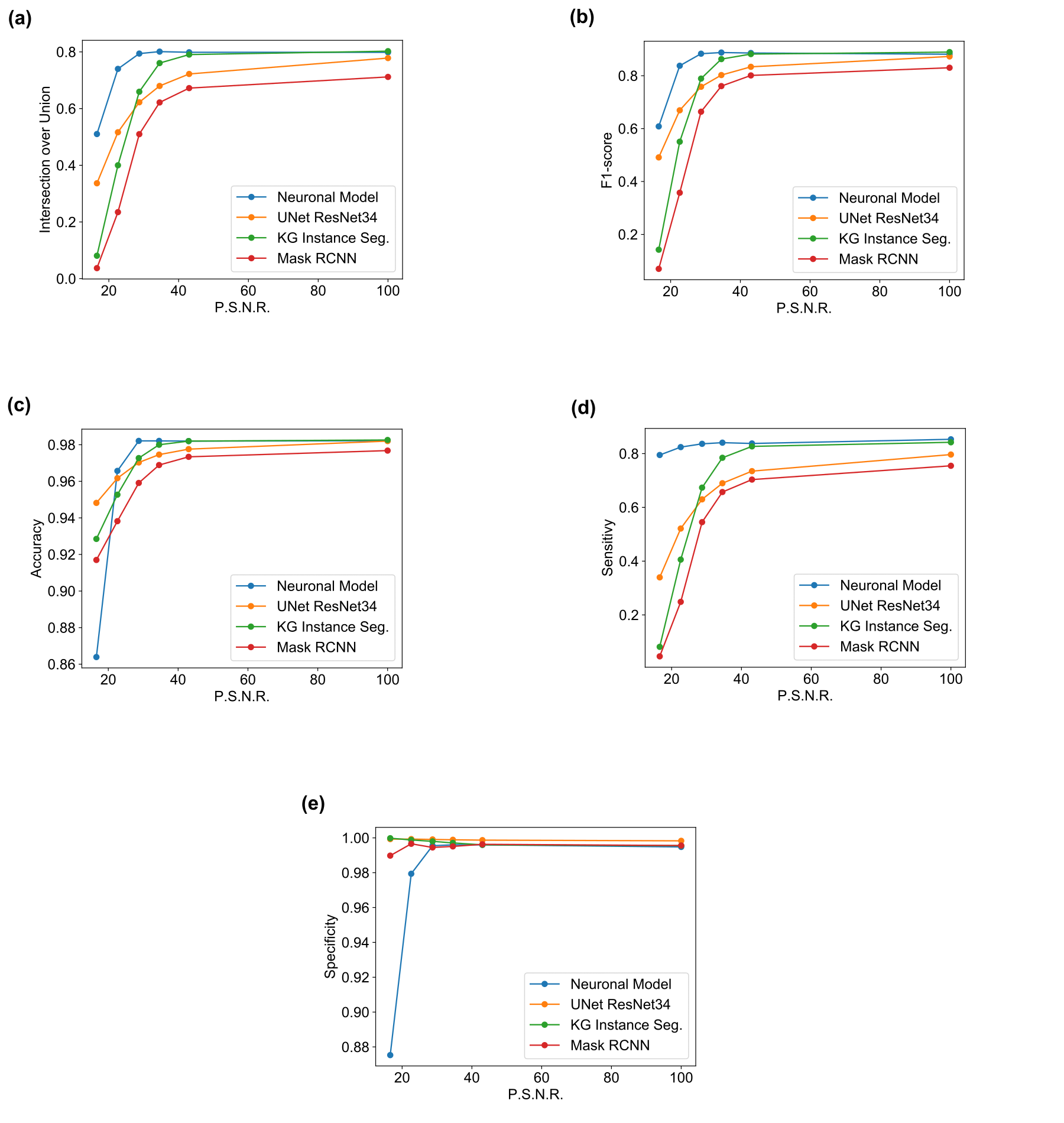}
	\caption{performances in terms of IoU (a), F1-score (b), Accuracy (c), Sensitivity (d) and Specificity (e) for each algorithm on NucleusSegData dataset.}
	\label{fig:fig7}
\end{figure}

\subsection{Results On Nucleussegdata Dataset}
Most of the time, the principal critique that moved towards non-deep-learning approaches is that they must be tuned on a single dataset (in worst cases on a single image) to achieve performances comparable to neural networks. Therefore, further tests were performed. This test attempts to quantify the generalization capability of the previously exposed algorithms. For example, the DNNs (U-Net, KG network, and R-CNN) segmented the images of the dataset NucleusSegData \cite{Gunesli2020} without fine-tuning. Such an effect could be a limitation; however, it is a typical pipeline in real-life applications. Indeed, in some cases, the DNN algorithm is connected to a camera that directly streams the image to the algorithm \cite{Bar2021}, and then the algorithm is applied to images that could differ substantially from the training and test sets used by the authors. In other cases, fine-tuning should follow strict rules \cite{AMIA}, which would make it impractical. Similarly, the neuronal model was used on the same dataset (NucleusSegData) without tuning any parameters. Table \ref{tab:t6} presents the performance of the four models with the NucleusSegData dataset.
Adversative noise was added to this dataset with P.S.N.R. 100.0, 43.0, 34.5, 28.7, 22.6, and 16.6. The results are presented in tables \ref{tab:t6}-\ref{tab:t10} (in the Appendix) and figure \ref{fig:fig7}. figures \ref{fig:fig7}a and \ref{fig:fig7}b with table \ref{tab:t6} and \ref{tab:t7} show how the proposed model outperformed the N.N.s for high-noise values. figure \ref{fig:fig7}c and table \ref{tab:t8} show that the proposed model has a high Accuracy for PSNR values greater than 16.6. However, for PSC =PSNR equal to 16.6, the Accuracy value decreases suddenly. This phenomenon is caused by the fact that the proposed model oversegments for high values of noise; instead, the other networks are more prone to undersegmentation.

\begin{figure}[h!]
	\centering
	\includegraphics[width=0.75\textwidth]{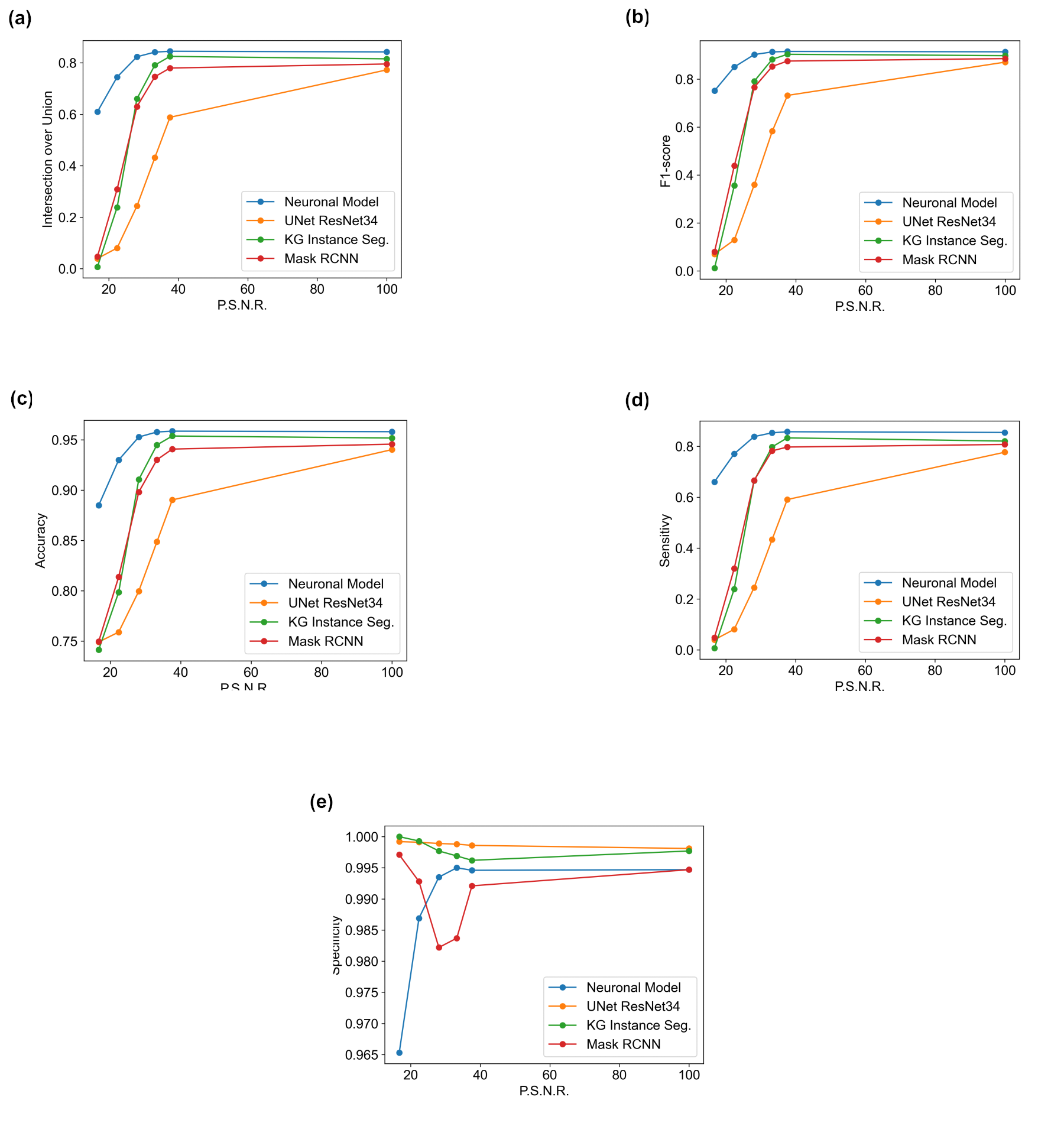}
	\caption{performances in terms of IoU (a), F1-score (b), Accuracy (c), Sensitivity (d) and Specificity (e) for each algorithm on ISBI 2009 dataset.}
	\label{fig:fig8}
\end{figure}

The same trend can be observed in the analysis of the ISBI dataset depicted in tables \ref{tab:t11}-\ref{tab:t15} (in the Appendix) and figures \ref{fig:fig8}a-e.

Although our method has formal correctness and high performance, it requires computational improvements because our method has a computation time approximately 10 times that of the DNN-proposed algorithms. However, in the defense of the method, it must be noted that the neural networks run on hardware optimized for this task (GPU). It has been found that the speedup of using a CPU instead of a GPU can reach a factor of 100 \cite{Lawrence2017}. However, our neuronal agents are fully parallel and run on a CPU, which opens the way for GPU implementation. To date, the most promising approach has been neuromorphic computing \cite{Rowley2018}. This approach could reduce the number of iterations required to compute the neuron model using a hardware circuit that reproduces neuron behavior, drastically reducing the time required. Such hardware has already been successfully tested on a multitude of tasks \cite{Orchard2021}.

\section{Conclusions} \label{sec:VI}
From a rough investigation of new methods in image analysis, it can be noted that there has been a constant explosion of techniques based on machine learning and its high level of flexibility. However, comprehension of these results is usually obscure. 
This evidence suggests that methodologies based on machine learning, deep learning, and general methods oriented toward artificial intelligence are the panacea for any problem present in the computational sciences. 
Nonetheless, many researchers continue to direct their research towards models that, on the one hand, do not involve the classic models of machine learning and, on the other hand, satisfy the performances of M.L. and their results are more comprehensive.
Therefore, some efforts have been made to formulate formal methods that are detached from any reference to artificial intelligence and are comparable to the most performing and current artificial intelligence techniques.
The planned contribution (NeuronalAlg) highlights a computational model based on deterministic foundations and not on a neuronal network style. Therefore, our primary purpose is to demonstrate its formal correctness, easy exploration, and comparable results to the best-performing A.I. models in the presence of adversative noise.
We showed good qualitative results, and a more accurate analysis was carried out with a set of quantitative indicators (see IoU, F1-Score, Accuracy, Specificity, and Sensitivity indexes in tables 1-10), which was necessary to compare them with the most common models of neuronal networks. The analysis performed on the original images and modified with different noise models demonstrated reliable and robust results. The graphs and tables extracted from the experiments indicate the unnecessary use of A.I. techniques, even when the application context requires formally correct models with demonstrable results.
Therefore, we propose a model that appears to be a valid alternative to N.N. methods in contexts where reliability and robustness must be formally verifiable, even for negligible percentages of error, which is necessary to understand the reasons for their occurrence.

\section*{Acknowledgments}
This research has received funding from the European Union’s Horizon 2020 Framework Programme for Research and Innovation under the Specific Grant Agreement numbers 945539 (Human Brain Project SGA3), Fenix computing and storage resources under the Specific Grant Agreement No. 800858 (Human Brain Project ICEI), and a grant from the Swiss National Supercomputing Centre (CSCS) under project ID ich002. Editorial support was provided by Annemieke Michels of the Human Brain Project. This work was partially financed by Project PO FESR Sicilia 2014/2020 - Azione 1.1.5. - "Sostegno all’avanzamento tecnologico delle imprese attraverso il finanziamento di linee pilota e azioni di validazione precoce dei prodotti e di dimostrazioni su larga scala" - 3DLab-Sicilia CUP G69J18001100007 - Number 08CT4669990220.

%Bibliography
\bibliographystyle{unsrt}  
\bibliography{arxiv_GGiacopelli} 

\newpage
\section*{Appendix}

\subsection*{Best, average and worst cases}
\begin{figure}[h!t]
	\begin{minipage}[b]{0.5\linewidth}
		\centering
		\includegraphics[width=0.65\textwidth]{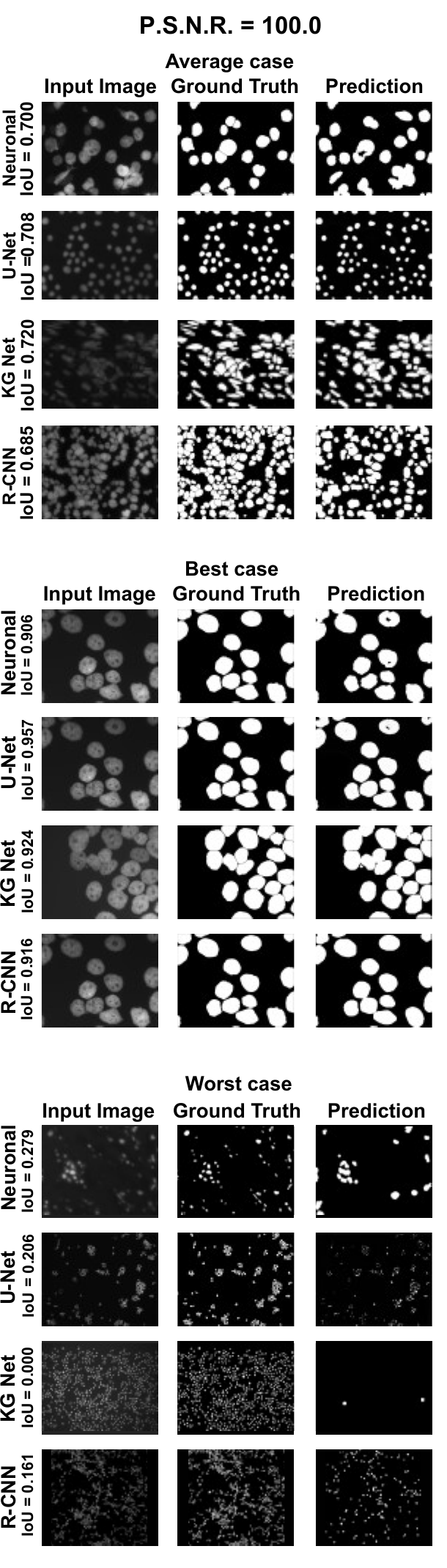}
		\caption{Average, best and worst case for each algorithm with PSNR = 100.0}
		\label{fig:fig9a}
	\end{minipage}
	\hspace{0.5cm}
	\begin{minipage}[b]{0.5\linewidth}
		\centering
		\includegraphics[width=0.65\textwidth]{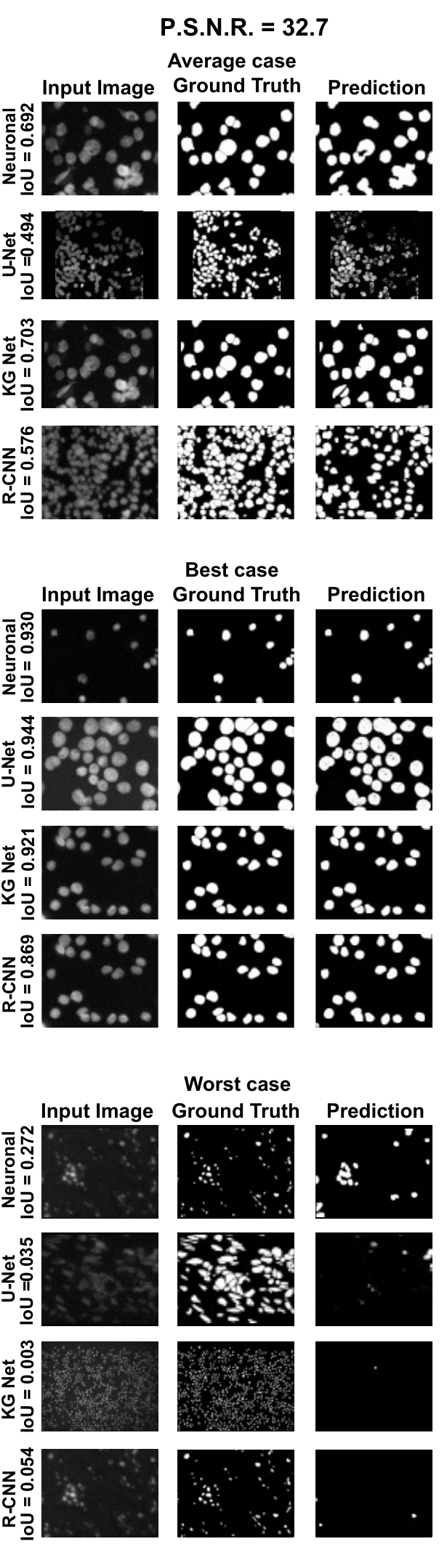}
		\caption{Average, best and worst case for each algorithm with P.S.N.R. = 32.7}
		\label{fig:fig9b}
	\end{minipage}
\end{figure}
\newpage

\begin{figure}[h!t]
	\begin{minipage}[b]{0.5\linewidth}
		\centering
		\includegraphics[width=0.65\textwidth]{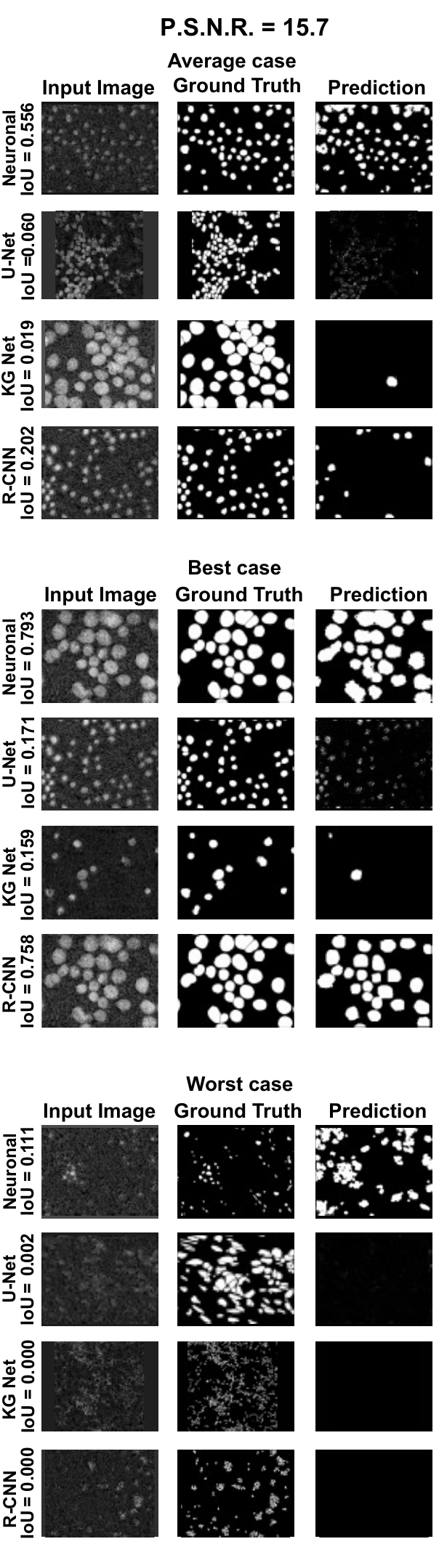}
		\caption{Average, best and worst case for each  algorithm with P.S.N.R. = 15.7}
		\label{fig:fig9c}
	\end{minipage}
	\hspace{0.5cm}
	\begin{minipage}[b]{0.5\linewidth}
		\centering
		\includegraphics[width=0.65\textwidth]{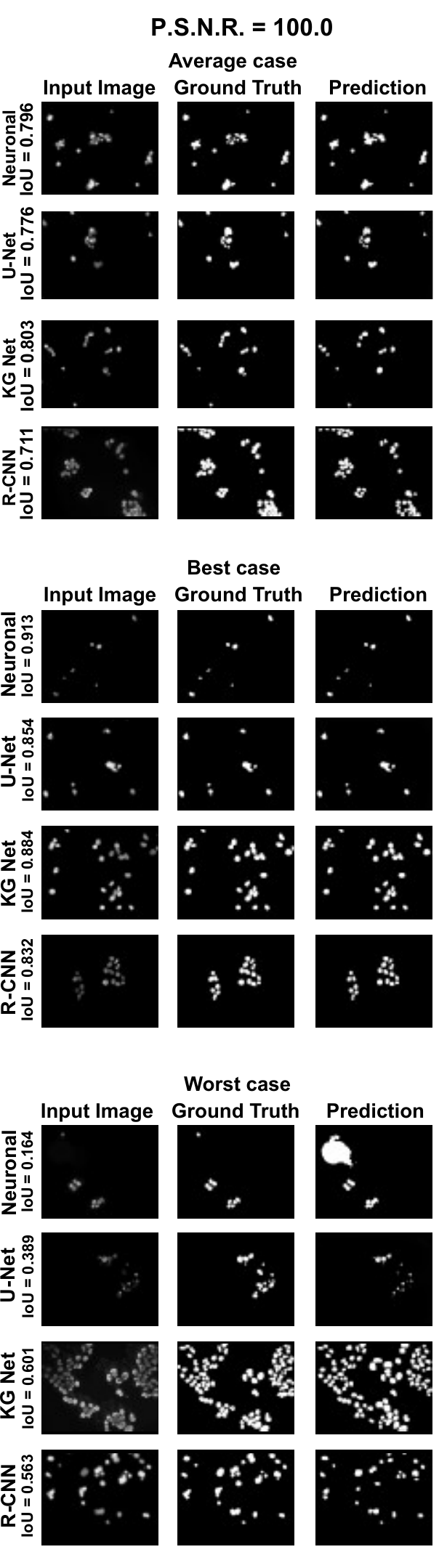}
		\caption{Average, best and worst case for each algorithm for the dataset NucleusSegData with P.S.N.R. = 100.0}
		\label{fig:fig10a}
	\end{minipage}
\end{figure}
\newpage

\begin{figure}[h!t]
	\begin{minipage}[b]{0.5\linewidth}
		\centering
		\includegraphics[width=0.65\textwidth]{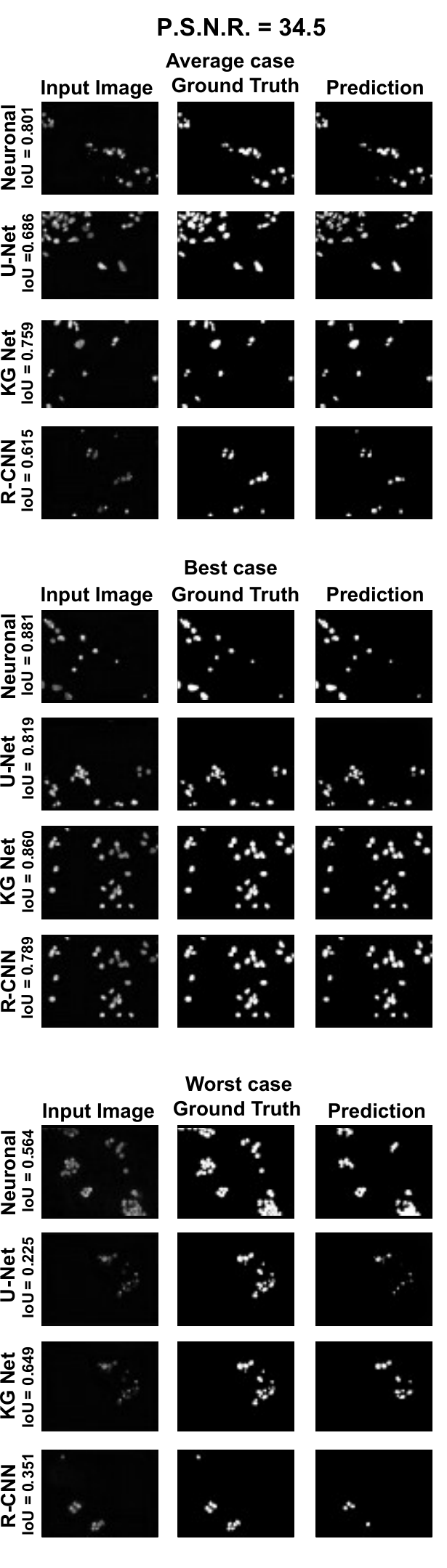}
		\caption{Average, best and worst case for each algorithm for the dataset NucleusSegData with P.S.N.R. = 34.5}
		\label{fig:fig10b}
	\end{minipage}
	\hspace{0.5cm}
	\begin{minipage}[b]{0.5\linewidth}
		\centering
		\includegraphics[width=0.65\textwidth]{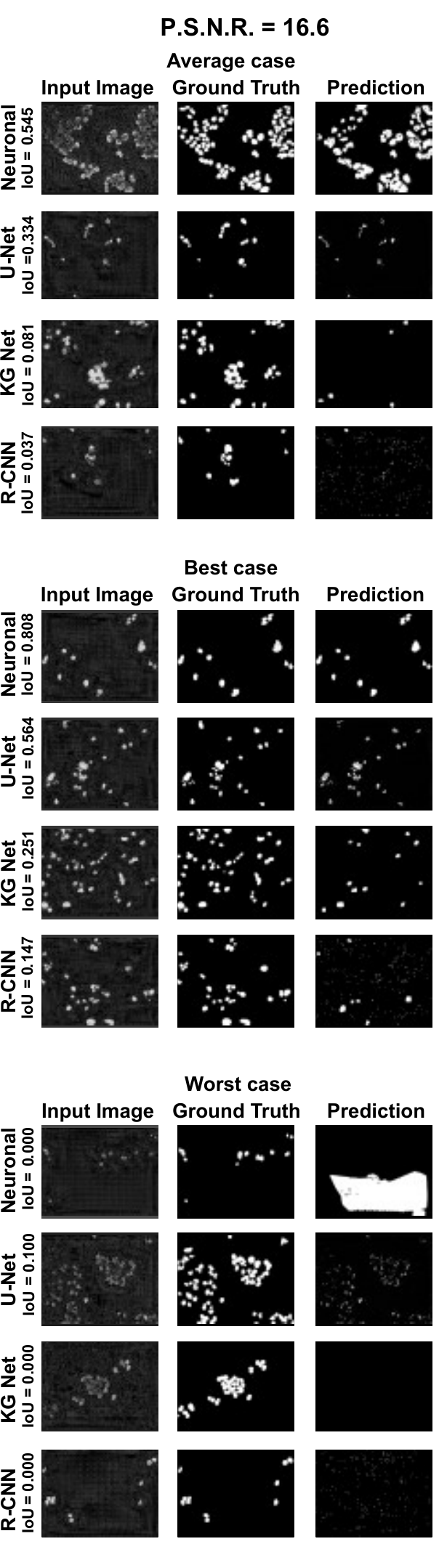}
		\caption{Average, best and worst case for each algorithm for the dataset NucleusSegData with P.S.N.R. = 16.6}
		\label{fig:fig10c}
	\end{minipage}
\end{figure}
\newpage

\begin{figure}[h!t]
	\begin{minipage}[b]{0.5\linewidth}
		\centering
		\includegraphics[width=0.65\textwidth]{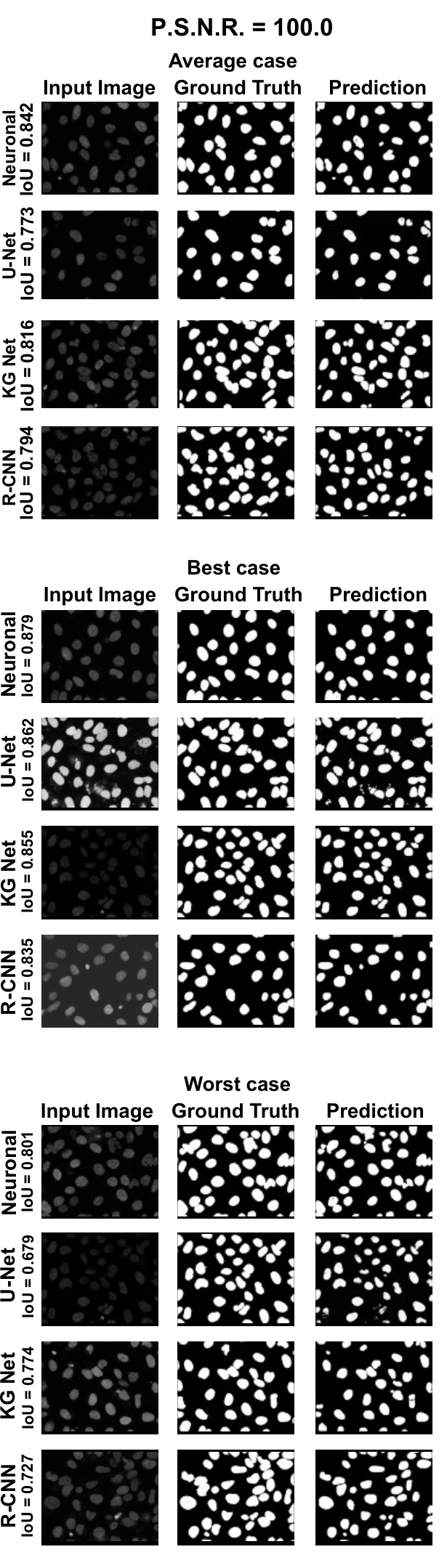}
		\caption{Average, best and worst case for each algorithm for the dataset ISBI 2009 with P.S.N.R. = 100.0}
		\label{fig:fig11a}
	\end{minipage}
	\hspace{0.5cm}
	\begin{minipage}[b]{0.5\linewidth}
		\centering
		\includegraphics[width=0.65\textwidth]{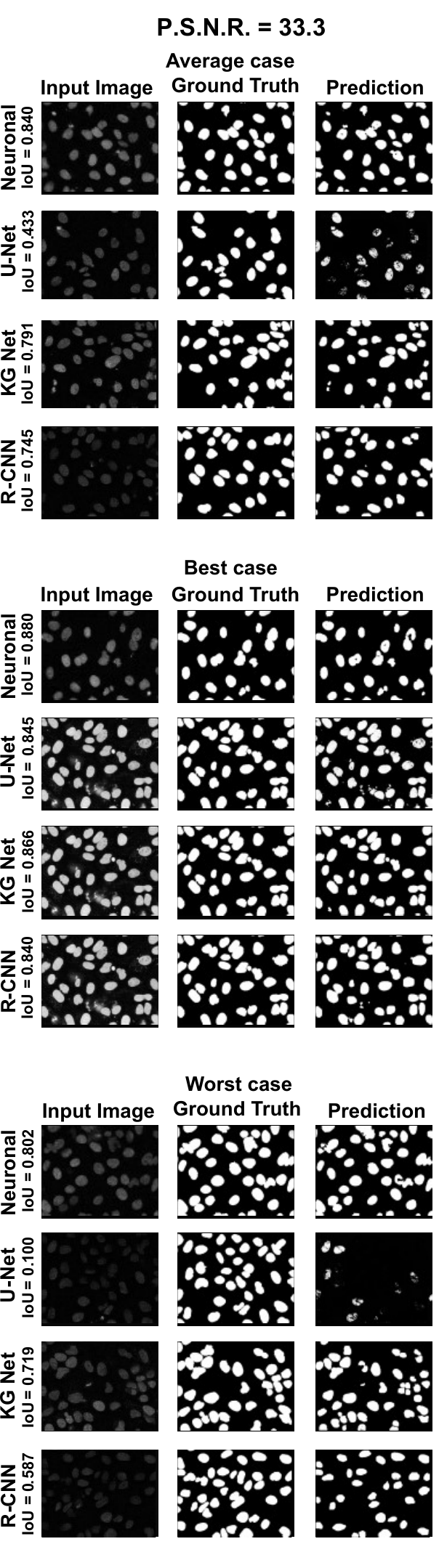}
		\caption{Average, best and worst case for each algorithm for the dataset ISBI 2009 with P.S.N.R. = 33.3}
		\label{fig:fig11b}
	\end{minipage}
\end{figure}
\newpage

\begin{figure}[h!t]
	\begin{minipage}[b]{0.5\linewidth}
		\centering
		\includegraphics[width=0.65\textwidth]{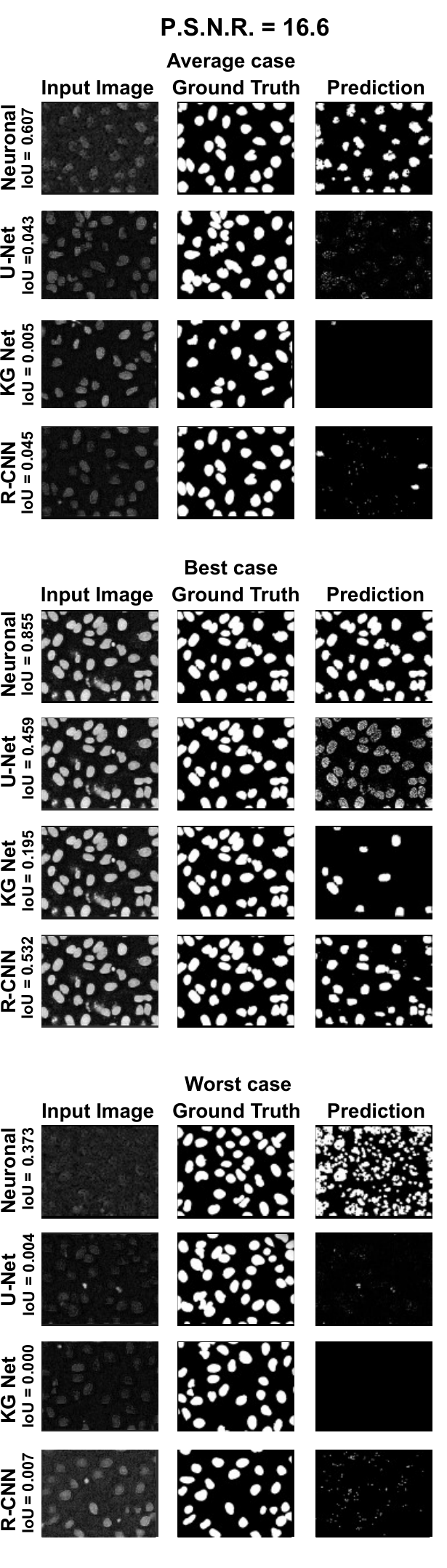}
		\caption{Average, best and worst case for each algorithm for the dataset ISBI 2009 with P.S.N.R. = 16.6}
		\label{fig:fig11c}
	\end{minipage}
	\hspace{0.5cm}
\end{figure}
\newpage

\newpage
\subsection*{Tables}
\begin{table}[h!]
	\caption{Intersection over Union values of the algorithms for different PSNR values of adversative noise (Neuroblastoma).}
	\centering
	\begin{tabular}{l|llllll}
		\toprule
		IoU & 100.0 & 40.1 & 32.7 & 26.9 & 21.1 & 15.7 \\
		\midrule
		Neuronal Alg. & 0.695 & 0.700 & 0.695 & 0.673 & 0.614 & 0.549 \\
		U-Net ResNet34 & 0.718 & 0.559 & 0.503 & 0.406 & 0.192 & 0.060 \\
		KG network & 0.712 & 0.704 & 0.664 & 0.593 & 0.319 & 0.025 \\
		Mask R-CNN & 0.682 & 0.645 & 0.576 & 0.489 & 0.348 & 0.207 \\
		\bottomrule
	\end{tabular}
	\label{tab:t1}
\end{table}

\begin{table}[h!]
	\caption{F1-score values of the algorithms for different PSNR values of adversative noise (Neuroblastoma).}
	\centering
	\begin{tabular}{l|llllll}
		\toprule
		F1-score & 100.0 & 40.1 & 32.7 & 26.9 & 21.1 & 15.7 \\
		\midrule
		Neuronal Alg. & 0.805 & 0.808 & 0.802 & 0.785 & 0.739 & 0.683 \\
		U-Net ResNet34 & 0.815 & 0.678 & 0.623 & 0.523 & 0.297 & 0.111 \\
		KG network & 0.796 & 0.787 & 0.754 & 0.688 & 0.435 & 0.047 \\
		Mask R-CNN & 0.787 & 0.755 & 0.695 & 0.606 & 0.457 & 0.287 \\
		\bottomrule
	\end{tabular}
	\label{tab:t2}
\end{table}

\begin{table}[h!]
	\caption{Accuracy values of the algorithms for different PSNR values of adversative noise (Neuroblastoma).}
	\centering
	\begin{tabular}{l|llllll}
		\toprule
		Accuracy & 100.0 & 40.1 & 32.7 & 26.9 & 21.1 & 15.7 \\
		\midrule
		Neuronal Alg. & 0.938 & 0.938 & 0.936 & 0.929 & 0.910 & 0.879 \\
		U-Net ResNet34 & 0.955 & 0.926 & 0.915 & 0.895 & 0.845 & 0.811 \\
		KG network & 0.953 & 0.952 & 0.945 & 0.932 & 0.873 & 0.804 \\
		Mask R-CNN & 0.943 & 0.936 & 0.923 & 0.908 & 0.881 & 0.849 \\
		\bottomrule
	\end{tabular}
	\label{tab:t3}
\end{table}

\begin{table}[h!]
	\caption{Sensitivity values of the algorithms for different PSNR values of adversative noise (Neuroblastoma).}
	\centering
	\begin{tabular}{l|llllll}
		\toprule
		Sensitivity & 100.0 & 40.1 & 32.7 & 26.9 & 21.1 & 15.7 \\
		\midrule
		Neuronal Alg. & 0.848 & 0.855 & 0.853 & 0.849 & 0.816 & 0.778 \\
		U-Net ResNet34 & 0.751 & 0.576 & 0.518 & 0.416 & 0.194 & 0.061 \\
		KG network & 0.801 & 0.779 & 0.732 & 0.637 & 0.331 & 0.026 \\
		Mask R-CNN & 0.751 & 0.699 & 0.626 & 0.530 & 0.377 & 0.221 \\
		\bottomrule
	\end{tabular}
	\label{tab:t4}
\end{table}

\begin{table}[h!]
	\caption{Specificity values of the algorithms for different PSNR values of adversative noise (Neuroblastoma).}
	\centering
	\begin{tabular}{l|llllll}
		\toprule
		Specificity & 100.0 & 40.1 & 32.7 & 26.9 & 21.1 & 15.7 \\
		\midrule
		Neuronal Alg. & 0.953 & 0.954 & 0.952 & 0.942 & 0.926 & 0.903 \\
		U-Net ResNet34 & 0.988 & 0.993 & 0.993 & 0.995 & 0.998 & 0.999 \\
		KG network & 0.973 & 0.976 & 0.976 & 0.981 & 0.991 & 0.9996 \\
		Mask R-CNN & 0.975 & 0.979 & 0.979 & 0.980 & 0.981 & 0.989 \\
		\bottomrule
	\end{tabular}
	\label{tab:t5}
\end{table}

\begin{table}[h!]
	\caption{Intersection over Union values of the algorithms for different PSNR values of adversative noise (NucleusSegData).}
	\centering
	\begin{tabular}{l|llllll}
		\toprule
		IoU & 100.0 & 43.0 & 34.5 & 28.7 & 22.6 & 16.6 \\
		\midrule
		Neuronal Alg. & 0.798 & 0.799 & 0.801 & 0.794 & 0.740 & 0.510 \\
		U-Net ResNet34 & 0.778 & 0.722 & 0.680 & 0.622 & 0.516 & 0.337 \\
		KG network & 0.802 & 0.790 & 0.760 & 0.659 & 0.400 & 0.081 \\
		Mask R-CNN & 0.712 & 0.672 & 0.622 & 0.510 & 0.235 & 0.037 \\
		\bottomrule
	\end{tabular}
	\label{tab:t6}
\end{table}

\begin{table}[h!]
	\caption{F1-score values of the algorithms for different PSNR values of adversative noise (NucleusSegData).}
	\centering
	\begin{tabular}{l|llllll}
		\toprule
		F1-score & 100.0 & 43.0 & 34.5 & 28.7 & 22.6 & 16.6 \\
		\midrule
		Neuronal Alg. & 0.881 & 0.886 & 0.888 & 0.883 & 0.838 & 0.608 \\
		U-Net ResNet34 & 0.873 & 0.834 & 0.802 & 0.758 & 0.669 & 0.491 \\
		KG network & 0.889 & 0.882 & 0.863 & 0.789 & 0.550 & 0.142 \\
		Mask R-CNN & 0.830 & 0.801 & 0.761 & 0.664 & 0.357 & 0.070 \\
		\bottomrule
	\end{tabular}
	\label{tab:t7}
\end{table}

\begin{table}[h!]
	\caption{Accuracy values of the algorithms for different PSNR values of adversative noise (NucleusSegData).}
	\centering
	\begin{tabular}{l|llllll}
		\toprule
		Accuracy & 100.0 & 43.0 & 34.5 & 28.7 & 22.6 & 16.6 \\
		\midrule
		Neuronal Alg. & 0.982 & 0.982 & 0.982 & 0.982 & 0.966 & 0.864 \\
		U-Net ResNet34 & 0.982 & 0.978 & 0.975 & 0.970 & 0.962 & 0.948 \\
		KG network & 0.983 & 0.982 & 0.980 & 0.973 & 0.953 & 0.928 \\
		Mask R-CNN & 0.977 & 0.973 & 0.969 & 0.959 & 0.938 & 0.917 \\
		\bottomrule
	\end{tabular}
	\label{tab:t8}
\end{table}

\begin{table}[h!]
	\caption{Sensitivity values of the algorithms for different PSNR values of adversative noise (NucleusSegData).}
	\centering
	\begin{tabular}{l|llllll}
		\toprule
		Sensitivity & 100.0 & 43.0 & 34.5 & 28.7 & 22.6 & 16.6 \\
		\midrule
		Neuronal Alg. & 0.853 & 0.838 & 0.841 & 0.836 & 0.824 & 0.795 \\
		U-Net ResNet34 & 0.796 & 0.735 & 0.690 & 0.630 & 0.521 & 0.340 \\
		KG network & 0.842 & 0.827 & 0.784 & 0.637 & 0.406 & 0.081 \\
		Mask R-CNN & 0.755 & 0.703 & 0.657 & 0.545 & 0.249 & 0.046 \\
		\bottomrule
	\end{tabular}
	\label{tab:t9}
\end{table}

\begin{table}[h!]
	\caption{Specificity values of the algorithms for different PSNR values of adversative noise (NucleusSegData).}
	\centering
	\begin{tabular}{l|llllll}
		\toprule
		Specificity & 100.0 & 43.0 & 34.5 & 28.7 & 22.6 & 16.6 \\
		\midrule
		Neuronal Alg. & 0.995 & 0.996 & 0.996 & 0.996 & 0.979 & 0.875 \\
		U-Net ResNet34 & 0.998 & 0.999 & 0.999 & 0.999 & 0.999 & 0.999 \\
		KG network & 0.996 & 0.996 & 0.997 & 0.998 & 0.999 & 1.0 \\
		Mask R-CNN & 0.996 & 0.996 & 0.995 & 0.995 & 0.997 & 0.990 \\
		\bottomrule
	\end{tabular}
	\label{tab:t10}
\end{table}

\begin{table}[h!]
	\caption{Intersection over Union values of the algorithms for different PSNR values of adversative noise (ISBI 2009).}
	\centering
	\begin{tabular}{l|llllll}
		\toprule
		IoU & 100.0 & 43.0 & 34.5 & 28.7 & 22.6 & 16.6 \\
		\midrule
		Neuronal Alg. & 0.842 & 0.845 & 0.841 & 0.823 & 0.744 & 0.610 \\
		U-Net ResNet34 & 0.773 & 0.588 & 0.431 & 0.244 & 0.080 & 0.039 \\
		KG network & 0.815 & 0.825 & 0.791 & 0.660 & 0.238 & 0.007 \\
		Mask R-CNN & 0.795 & 0.780 & 0.746 & 0.630 & 0.308 & 0.047 \\
		\bottomrule
	\end{tabular}
	\label{tab:t11}
\end{table}

\begin{table}[h!]
	\caption{F1-score values of the algorithms for different PSNR values of adversative noise (ISBI 2009).}
	\centering
	\begin{tabular}{l|llllll}
		\toprule
		F1-score & 100.0 & 43.0 & 34.5 & 28.7 & 22.6 & 16.6 \\
		\midrule
		Neuronal Alg. & 0.914 & 0.916 & 0.914 & 0.903 & 0.851 & 0.752 \\
		U-Net ResNet34 & 0.871 & 0.732 & 0.583 & 0.360 & 0.130 & 0.070 \\
		KG network & 0.898 & 0.904 & 0.883 & 0.792 & 0.356 & 0.012 \\
		Mask R-CNN & 0.886 & 0.876 & 0.853 & 0.766 & 0.439 & 0.080 \\
		\bottomrule
	\end{tabular}
	\label{tab:t12}
\end{table}

\begin{table}[h!]
	\caption{Accuracy values of the algorithms for different PSNR values of adversative noise (ISBI 2009).}
	\centering
	\begin{tabular}{l|llllll}
		\toprule
		Accuracy & 100.0 & 43.0 & 34.5 & 28.7 & 22.6 & 16.6 \\
		\midrule
		Neuronal Alg. & 0.958 & 0.959 & 0.958 & 0.953 & 0.930 & 0.885 \\
		U-Net ResNet34 & 0.940 & 0.890 & 0.849 & 0.800 & 0.759 & 0.749 \\
		KG network & 0.952 & 0.954 & 0.945 & 0.911 & 0.799 & 0.741 \\
		Mask R-CNN & 0.946 & 0.941 & 0.930 & 0.898 & 0.814 & 0.750 \\
		\bottomrule
	\end{tabular}
	\label{tab:t13}
\end{table}

\begin{table}[h!]
	\caption{Sensitivity values of the algorithms for different PSNR values of adversative noise (ISBI 2009).}
	\centering
	\begin{tabular}{l|llllll}
		\toprule
		Sensitivity & 100.0 & 43.0 & 34.5 & 28.7 & 22.6 & 16.6 \\
		\midrule
		Neuronal Alg. & 0.854 & 0.857 & 0.853 & 0.838 & 0.770 & 0.660 \\
		U-Net ResNet34 & 0.777 & 0.591 & 0.433 & 0.245 & 0.081 & 0.040 \\
		KG network & 0.821 & 0.833 & 0.798 & 0.665 & 0.239 & 0.007 \\
		Mask R-CNN & 0.807 & 0.797 & 0.782 & 0.666 & 0.320 & 0.048 \\
		\bottomrule
	\end{tabular}
	\label{tab:t14}
\end{table}

\begin{table}[h!]
	\caption{Specificity values of the algorithms for different PSNR values of adversative noise (ISBI 2009).}
	\centering
	\begin{tabular}{l|llllll}
		\toprule
		Specificity & 100.0 & 43.0 & 34.5 & 28.7 & 22.6 & 16.6 \\
		\midrule
		Neuronal Alg. & 0.995 & 0.995 & 0.995 & 0.994 & 0.987 & 0.965 \\
		U-Net ResNet34 & 0.998 & 0.999 & 0.999 & 0.999 & 0.999 & 0.999 \\
		KG network & 0.998 & 0.996 & 0.997 & 0.998 & 0.999 & 1.0 \\
		Mask R-CNN & 0.995 & 0.992 & 0.984 & 0.982 & 0.993 & 0.997 \\
		\bottomrule
	\end{tabular}
	\label{tab:t15}
\end{table}

\end{document}